\documentclass[10pt,twocolumn,letterpaper]{article}

\usepackage{times}
\usepackage{graphicx}
\usepackage{amsmath}
\usepackage{amssymb}
\usepackage{booktabs}
\usepackage{caption}
\usepackage{subcaption}
\usepackage{authblk}
\usepackage{cite}
\usepackage{url}
\usepackage{enumitem}
\usepackage{listings}
\usepackage{xcolor}
\usepackage{svg}
\usepackage{float}

\definecolor{codegray}{rgb}{0.95,0.95,0.95}

\usepackage{tikz}
\usetikzlibrary{positioning, shadows, shapes.geometric, backgrounds}
\usepackage{fontawesome}
\usepackage[margin=1in]{geometry}

\lstdefinestyle{glsl}{
backgroundcolor=\color{codegray},
  language=C++,                    % Closest to GLSL syntax
  basicstyle=\ttfamily\small,
  keywordstyle=\color{blue},
  commentstyle=\color{gray},
  stringstyle=\color{orange},
  showstringspaces=false,
  breaklines=true,
  frame=single,
  captionpos=b,
  tabsize=2,
  numbers=none,
  numberstyle=\tiny\color{gray},
}
\title{Part Segmentation of Human Meshes \\ via Multi-View Human Parsing}

\author[1]{James Dickens}
\author[2]{Kamyar Hamad}
\affil[1]{Department of Computer Science, University of Ottawa}
\affil[2]{Department of Automation, Central South University}
\affil[ ]{\texttt{jdick088@uottawa.ca, kamyarothman@csu.edu.cn}}

\date{}

\begin{document}

\maketitle

\begin{abstract}
\textit{Recent advances in point cloud deep learning have led to models that achieve high per-part labeling accuracy on large-scale point clouds, using only the raw geometry of unordered point sets. In parallel, the field of human parsing focuses on predicting body part and clothing/accessory labels from images. This work aims to bridge these two domains by enabling per-vertex semantic segmentation of large-scale human meshes. To achieve this, a pseudo-ground truth labeling pipeline is developed for the Thuman2.1 dataset: meshes are first aligned to a canonical pose, segmented from multiple viewpoints, and the resulting point-level labels are then backprojected onto the original mesh to produce per-point pseudo ground truth annotations. Subsequently, a novel, memory-efficient sampling strategy is introduced—windowed iterative farthest point sampling (FPS) with space-filling curve-based serialization—to effectively downsample the point clouds. This is followed by a purely geometric segmentation using PointTransformer, enabling semantic parsing of human meshes without relying on texture information. Experimental results confirm the effectiveness and accuracy of the proposed approach. Project code and pre-processed data is available at \url{https://github.com/JamesMcCullochDickens/Human3DParsing/tree/master}.}
\end{abstract}

\section{Introduction}
High-quality part segmentation of 3D human meshes is a useful tool for applications such as character animation and game development, where fine-grained control over character models is desired. While many publicly available 3D models include skeletal rigs—hierarchies of anatomical keypoints used for animation—they often lack per-vertex part labels. Additionally, some models have incomplete or missing texture maps, making it difficult to apply color-based 2D-to-3D segmentation approaches.
\par To address this gap, in this work a pipeline that automatically generates 3D parsing labels from textured 3D human models, and trains a deep neural network to predict these labels using only geometric information is developed. Unlike existing approaches that rely on synthetic datasets where human instances are fit to parametric mesh models such as SMPL or SMPL-X \cite{SMPL:2015, SMPL-X:2019}, our method operates directly on raw, real-world 3D meshes obtained from a multi-view camera setup, which often exhibit greater diversity in body shapes, clothing, and poses.
\par This work draws inspiration from two complementary areas: 2D human parsing and 3D deep learning. The former refers to the task of segmenting human body parts and clothing in color images \cite{liang2015deep}, a well-studied domain with a rich set of pretrained models. These models will be leveraged to project 2D parsing labels into 3D space via multi-view backprojection and aggregation, enabling the creation of pseudo ground truth for 3D mesh segmentation.
\par The latter area, known as 3D deep learning, has emerged as a first-class research area in modern computer vision, focused largely on learning representations from point clouds and polygonal meshes. In the proposed approach, the vertices of polygonal meshes are treated as point clouds, wherein 3D deep learning techniques for semantic segmentation can be applied. However, existing point cloud models are typically designed for smaller point clouds (e.g., 2048 points), while human meshes can contain millions of vertices with redundant and densely packed regions. To address this, an efficient point cloud downsampling strategy that preserves semantic structure for training is introduced, followed by a simple upsampling stage to produce full-resolution mesh segmentations. 
\par Further, many 3D models do not come in a cannonical orientation, wherein human parsing models are most accurate when individuals face the camera in commonly encountered poses, i.e. front facing with minimal self-occlusions. To solve this issue, a keypoint-based approach locates anatomical points of interest, where rotations can be employed to correct a wide range of less desirable poses. In summary, the main contributions of this work are as follows:
\begin{itemize}
\item We propose a pipeline for generating pseudo ground truth 3D parsing labels by aggregating multi-view projections from 2D human parsing models, including an alignment step making use of keypoint-based correction of the input orientation of the 3D model.
\item We develop a memory-efficient point cloud sampling and upsampling strategy to enable full-resolution part segmentation of high-density human meshes.
\end{itemize}
\section{Related Work}
\subsection{Human Parsing}
One of the earliest works with respect to human parsing was developed by Yamaguchi et al. \cite{yamaguchi2012parsing}, in which the Fashionista dataset was introduced, using body parts as well as clothing and accessory labels to supervise a segmentation algorithm based on superpixels in a conditional random field framework, additionally making use of keypoints for refinement.
\par An early deep learning method by Liang et al. introduced the Active Template Regression (ATR) model \cite{liang2015deep}, which employed dictionary learning for each semantic label's corresponding part mask. The model used two convolutional neural networks (CNNs) to predict dictionary coefficients for part masks and estimated shape parameters. These predictions were fused through interpolation and further refined using superpixel segmentation.
\par Human parsing models are broadly categorized into two main approaches. In the bottom-up approach, individual parts are first detected and segmented, followed by a grouping stage that assembles them into person instances. An example of this is the Parsing Group Network (PGN) proposed by Gong et al. \cite{gong2018instance}, which employs a semantic segmentation branch to predict part instances and an edge detection branch to identify instance boundaries. The final parsing result is obtained by combining semantic part labels with instance edge information for effective grouping. 
\par By contrast, the top-down approach begins by detecting individual human instances using a person detection algorithm, often combined with instance segmentation to distinguish overlapping individuals. Semantic segmentation is then applied within each detected region, as exemplified by Parsing R-CNN \cite{yang2019parsingrcnn}, a two-stage framework. Single stage top-down approaches do not employ region proposal networks, such as NanoParsing \cite{xu2022nanoparsing}.
\par With respect to more recent approaches, a high-performing modern approach is the Mask2Former Parsing (M2FP) model \cite{yang2023humanparsing} developed by Liang et al., which extends the MaskFormer universal image segmentation algorithm of Kirillov et al. \cite{cheng2021maskformer, cheng2021mask2former} by predicting background queries, as well as part and person queries. More recently, the Sapiens model developed by Khirodkar et al \cite{khirodkar2024sapiens} introduces a foundation model, pre-trained using a masked auto-encoder (MAE) framework on the high resolution proprietary dataset Humans-300M. The resulting model is fine-tuned for human parsing using 28 part labels, expanding previous label categories to fine-grained facial attributes such as eyes, nose, teeth, lips, in addition to a distinction between arms and hand label categories.  
\par Publicly available datasets for human parsing are relatively small in terms of scale, and have conflicting label spaces. Most notably, the CIHP (crowd instance human parsing) \cite{gong2018instance}, LIP (look into person)\cite{gong2017look}, MHP-v2 (multi-human parsing) \cite{zhao2018understandingMHVP2} and Pascal Person-Part \cite{chen2014detect} datasets are commonly used for benchmarking, where their train/test sizes and number of labels is shown in Table 2.1. 
\begin{table}[ht]
\centering
\label{tab:dataset_stats}
\begin{tabular}{lcc}
\toprule
\textbf{Dataset} & \textbf{Train/Test} & \textbf{\# Parsing Labels} \\
\midrule
PASCAL Person-Part & 1,716/1,817 & 6 \\
MHP v2.0 & 15,403/5,000 & 58 \\
CIHP & 28,280/5,000 & 20 \\
LIP & 30,462/10,000 & 19 \\
\bottomrule
\end{tabular}
\caption*{\textit{Table 2.1: The number of train/test images for the most commonly used human publicly available parsing datasets in the literature, alongside the number of parsing labels considered respectively.}}
\end{table}

\begin{figure*}[t]
    \centering
   \def\svgwidth{0.85\textwidth}
    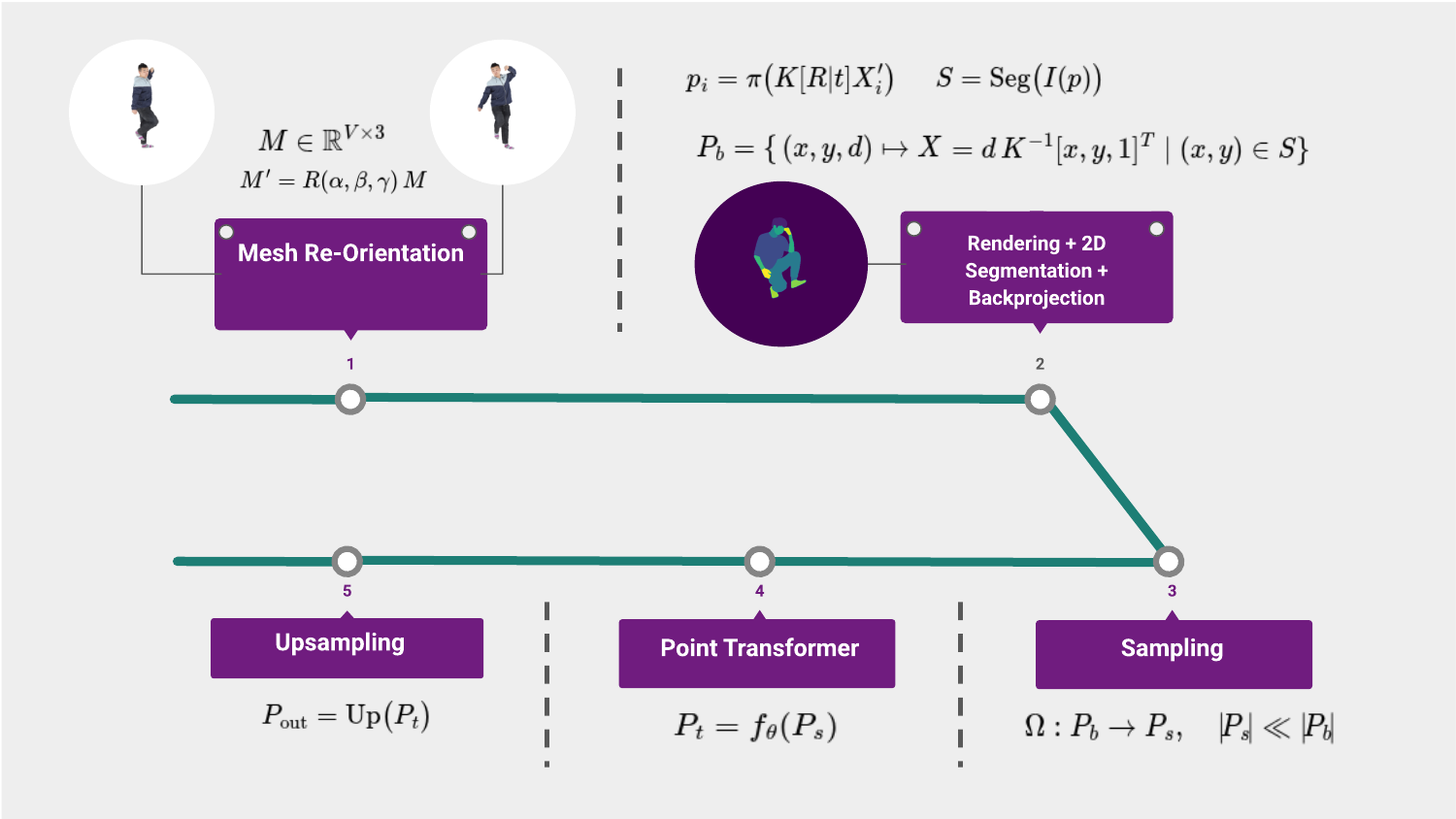
    \caption*{\textit{Figure 3.1, an overview of the proposed approach for parsing of human mesh models. Initially a mesh $M$ is aligned to a canonical orientation. From a set of views $v_i$, the meshes are rendered and segmented with a human parsing model, where the resulting point cloud is downsampled with the proposed windowed iterative farthest point algorithm, obtaining pseudo ground truth labels on a dataset. A point transformer model is trained to predict per-point labels, followed by nearest neighbors upsampling to the full mesh vertex resolution.}}
    \label{fig:orientation-adjust}
\end{figure*}

\subsection{Point Cloud Deep Learning and Semantic Segmentation}
As noted by Lu et al. \cite{lu2022transformers}, the bulk of modern point
cloud deep learning has largely focused on point cloud classification,
part segmentation, semantic segmentation, and 3D object
detection and tracking.

\par The seminal work of PointNet \cite{qi2017pointnet} introduced by Qi et al.
developed a set-based approach to point cloud classification
and part-segmentation. Points are mapped to a canonical
embedding space with multiple layers of learnable T-Net
(Transformation Net) operations, using max pooling as a set
aggregation tool, with individual point-wise features learned
by multi-layer perceptrons (MLPs). The follow-up work
PointNet++ \cite{qi2017pointnetplusplus} introduced the novel use of neighborhood
grouping by radius search for local feature aggregation, in
addition to the use of multi-scale (in terms of points) and
multi-radius feature learning, where points are successively
downsampled along the network's depth according to the iterative
farthest point sampling algorithm (IFPS). Another popular early work, Dynamic Graph
CNN \cite{wang2019dynamic} constructs dynamic K-NN graphs based on spatial
proximity alongside point-wise features of neighborhood
points for local feature aggregation.
\par In keeping with the trend in modern computer vision
to explore the use of concepts from the Transformer architecture \cite{vaswani2017attention}, PointTransformer v1 \cite{zhao2021point} employs
self-attention to aggregate features in local
radii around points, using relative positional embeddings,
employed for both point cloud classification and segmentation. The follow-up model PointTransformer v2 \cite{wu2022point} utilized
channel groupings in the attention mechanism, and uniform
grid-based pooling in the down-sampling and upsampling
stages of the network. Recently, PointTransformer v3 \cite{zhao2023point}
focuses on large-scale semantic segmentation by using
point serialization with space-filling curves (z-order curves
and Hilbert curves), to learn features within windows of 1
dimensional arrays of points, introducing serialized pooling/unpooling. They leverage serialization to avoid memory-expensive neighborhood searches.

\begin{figure*}[h!]
    \centering
    \resizebox{0.75\textwidth}{!}{%
    \begin{tikzpicture}[
        node distance=0.8cm and 0.5cm,
        auto,
        image node/.style={
            inner sep=0pt,
            drop shadow={shadow scale=1.05, shadow xshift=2pt, shadow yshift=-2pt, opacity=0.3}
        },
        arrow/.style={
            ->, 
            thick, 
            blue!70!black,
            shorten >=2pt,
            shorten <=2pt
        },
        label/.style={font=\small, text=black},
        iteration label/.style={
            font=\scriptsize\bfseries,
            text=white,
            fill=blue!70!black,
            rounded corners=2pt,
            inner sep=2pt,
            minimum width=1.2cm
        },
        step label/.style={
            font=\footnotesize,
            text=gray!80!black,
            align=center
        }
    ]
    
        % Top row
        \node[image node] (step0) {\includegraphics[width=2.6cm]{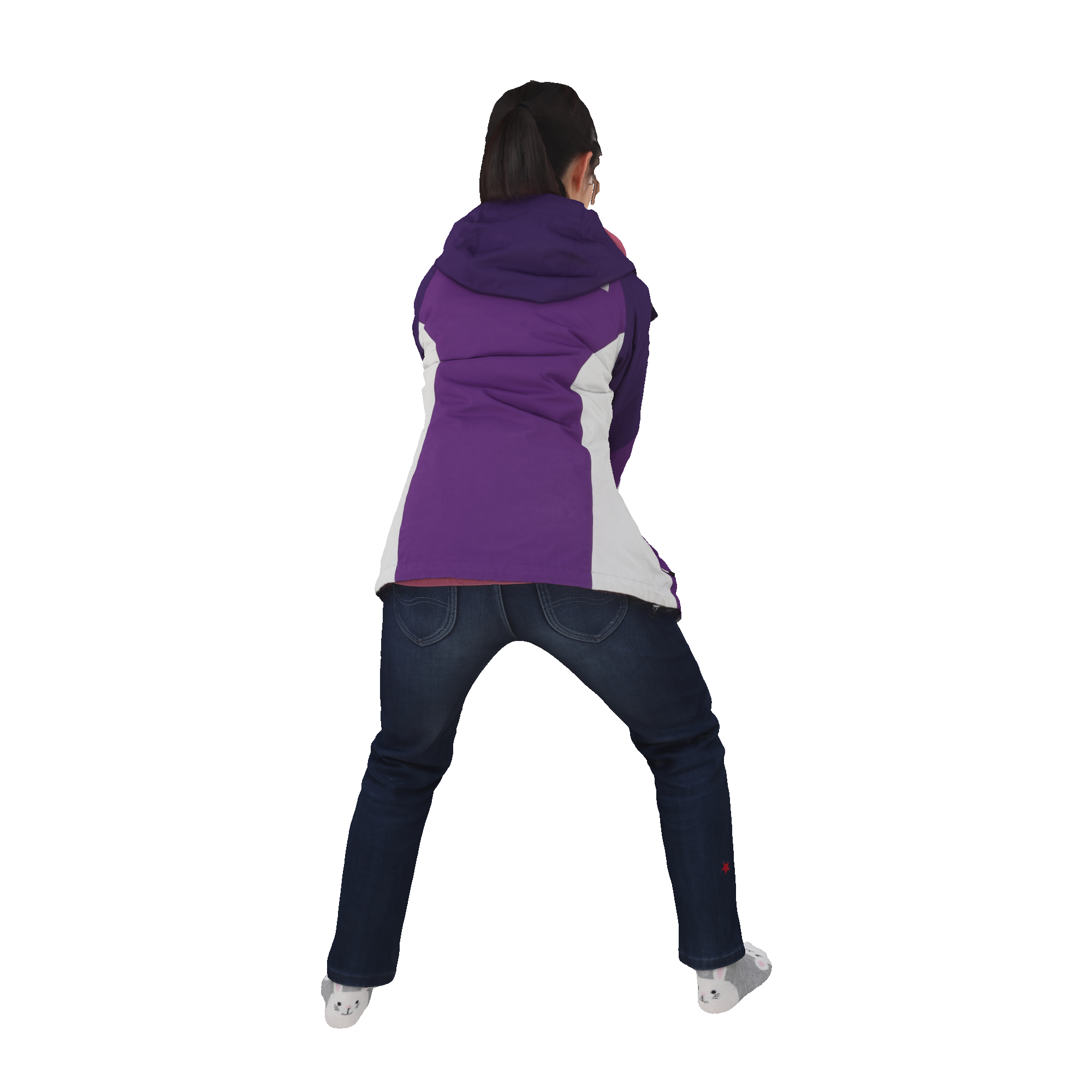}};
        \node[image node, right=of step0] (step1) {\includegraphics[width=2.6cm]{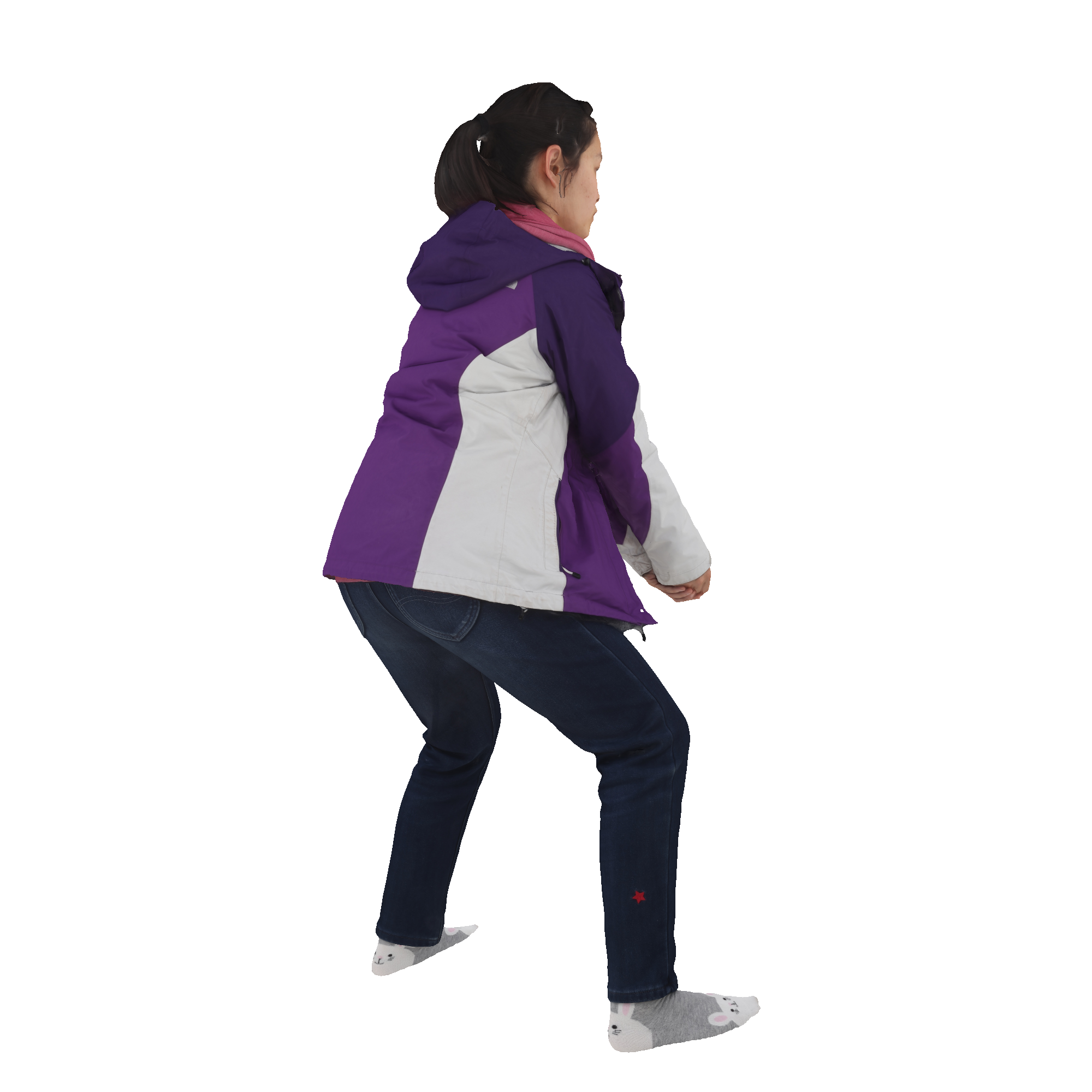}};
        \node[image node, right=of step1] (step2) {\includegraphics[width=2.6cm]{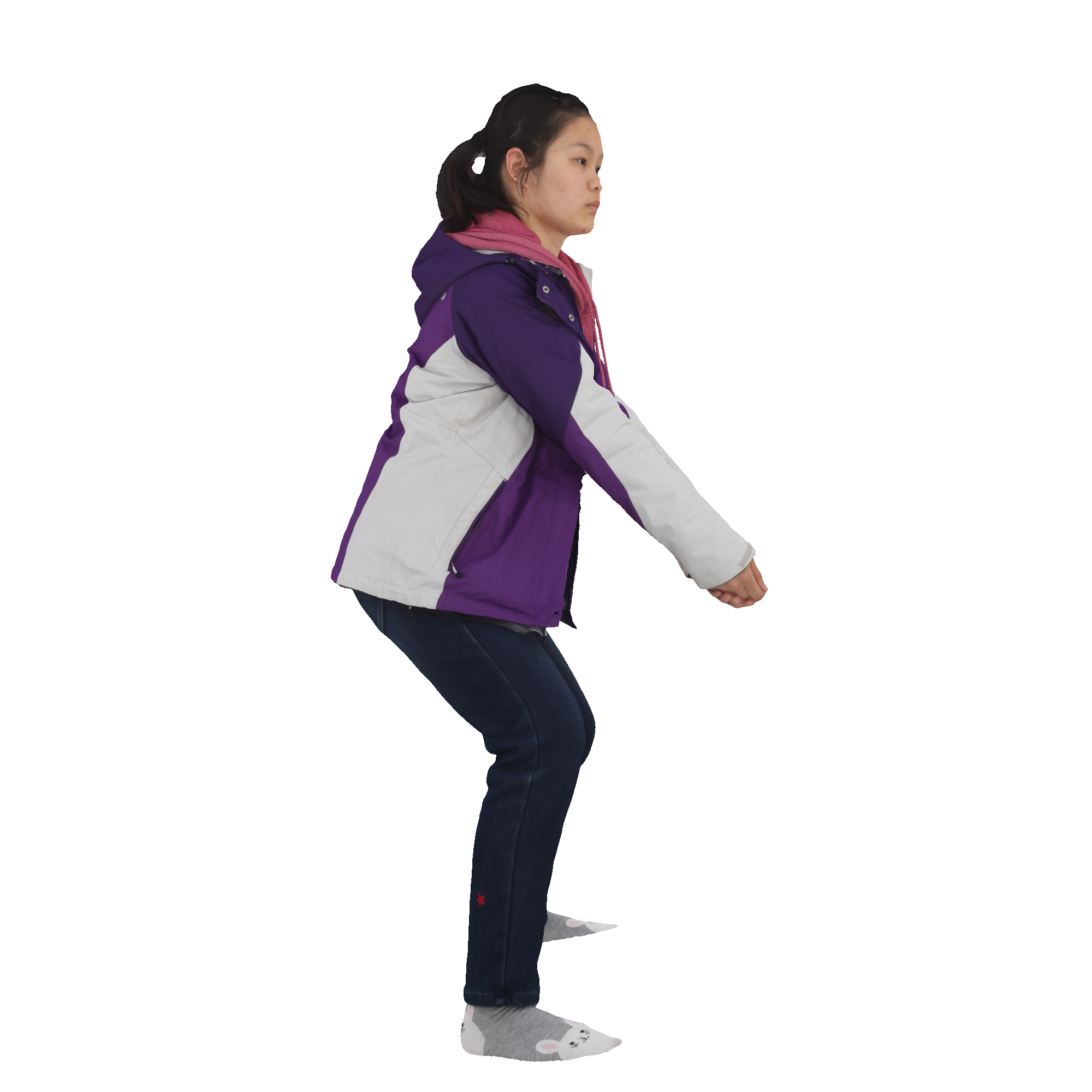}};
        \node[image node, right=of step2] (step3) {\includegraphics[width=2.6cm]{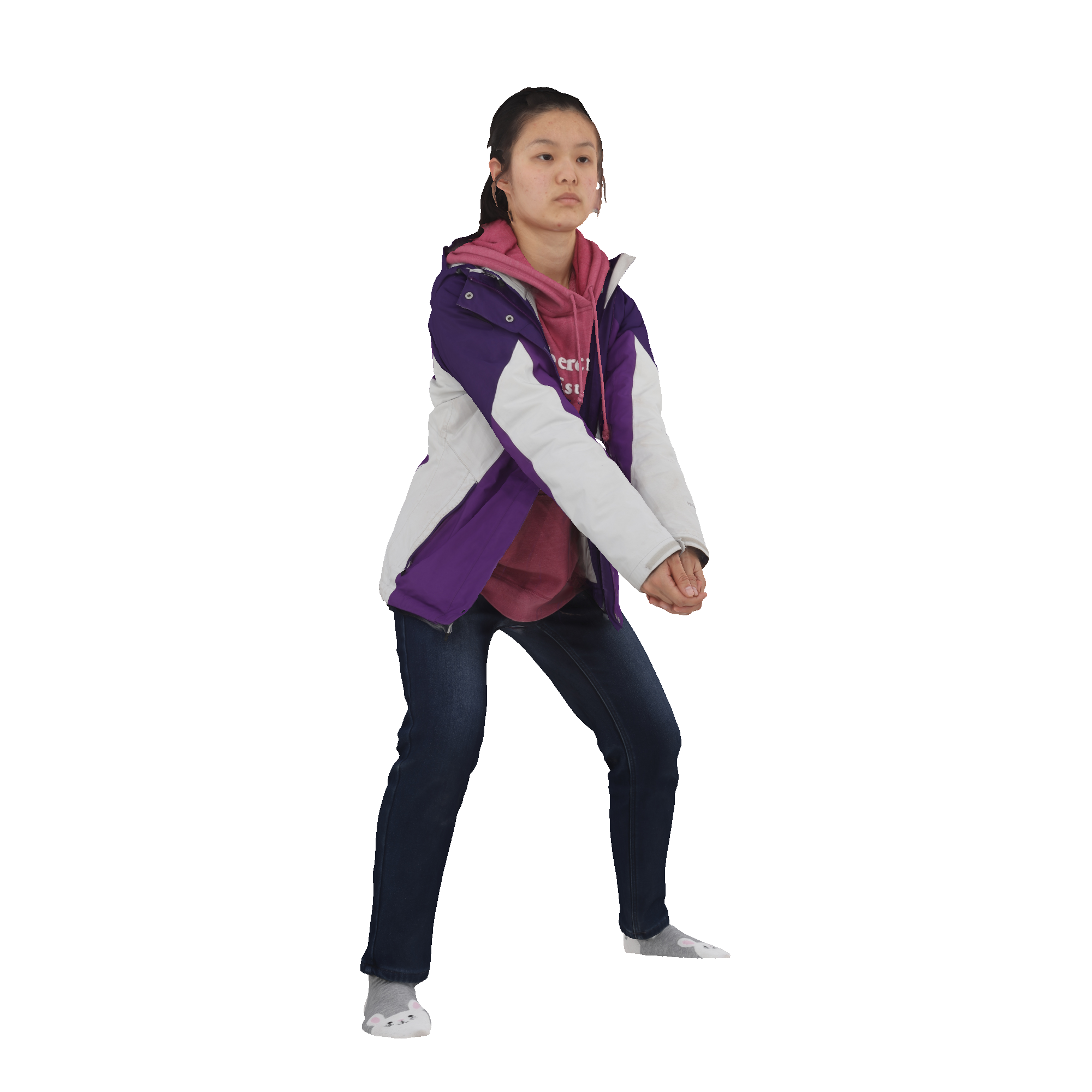}};
        \node[image node, right=of step3] (step4) {\includegraphics[width=2.6cm]{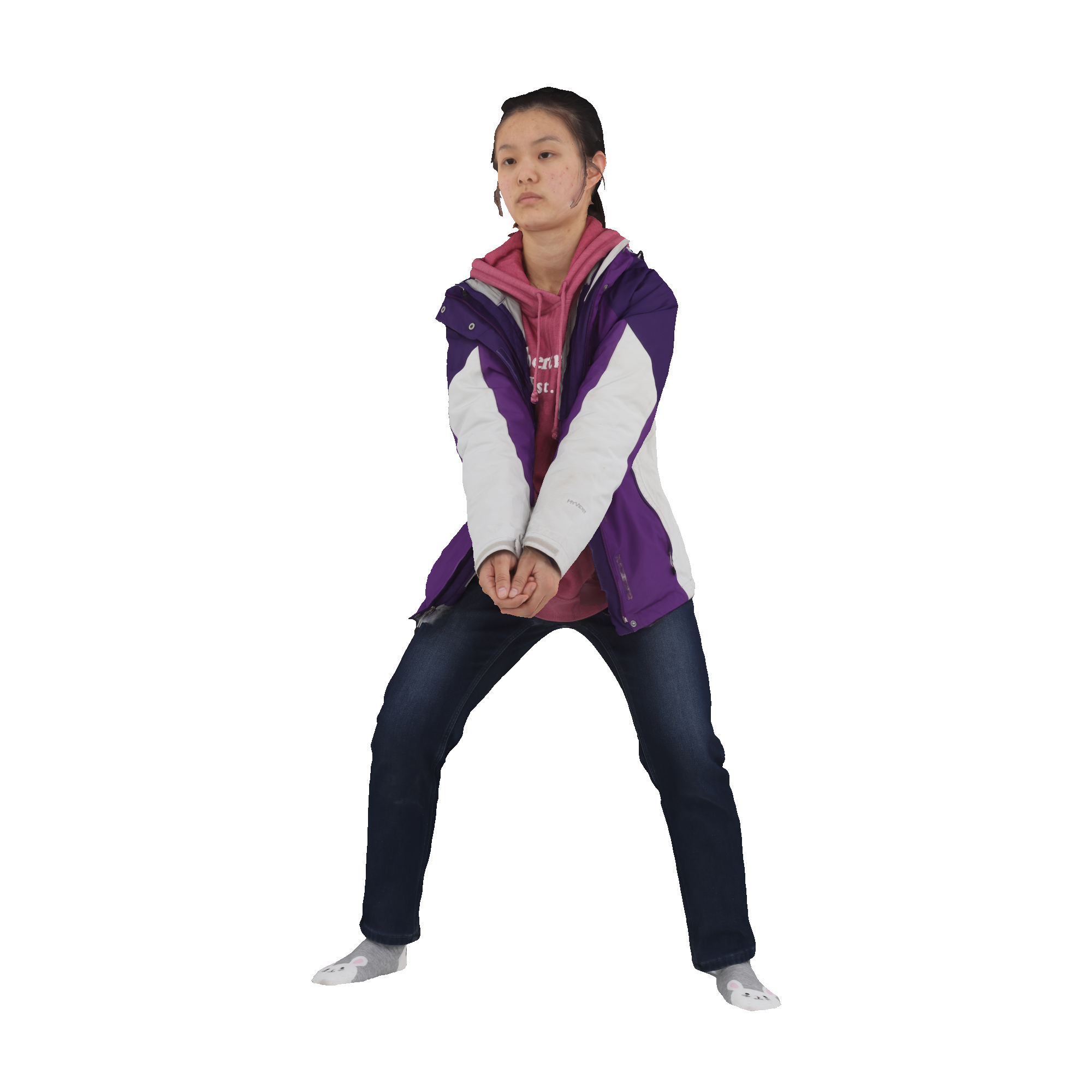}};
        
        \draw[arrow] (step0) -- (step1) node[midway, above=3pt, iteration label] {Iter 1};
        \draw[arrow] (step1) -- (step2) node[midway, above=3pt, iteration label] {Iter 2};
        \draw[arrow] (step2) -- (step3) node[midway, above=3pt, iteration label] {Iter 3};
        \draw[arrow] (step3) -- (step4) node[midway, above=3pt, iteration label] {Final};
        
        \node[below=0.25cm of step0, step label] {Initial\\Model};
        \node[below=0.25cm of step1, step label] {Adjusted\\Step 1};
        \node[below=0.25cm of step2, step label] {Adjusted\\Step 2};
        \node[below=0.25cm of step3, step label] {Adjusted\\Step 3};
        \node[below=0.25cm of step4, step label] {Final\\Orientation};

        \draw[gray!40, thick, dashed] 
            ([yshift=-0.8cm]step0.south) -- 
            ([yshift=-0.8cm]step4.south);

        % Bottom row
        \node[image node, below=1.5cm of step0] (alt0) {\includegraphics[width=3.2cm]{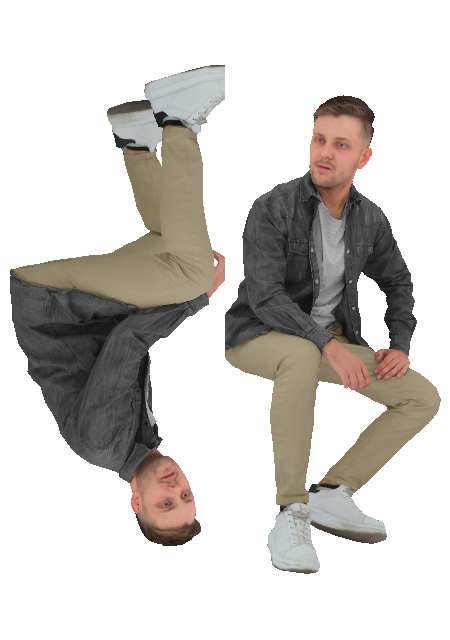}};
        \node[image node, right=1.0cm of alt0] (alt1) {\includegraphics[width=3.2cm]{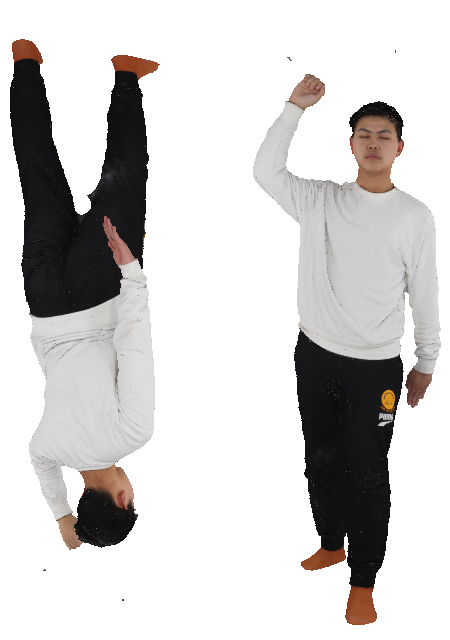}};
        \node[image node, right=1.0cm of alt1] (alt2) {\includegraphics[width=3.2cm]{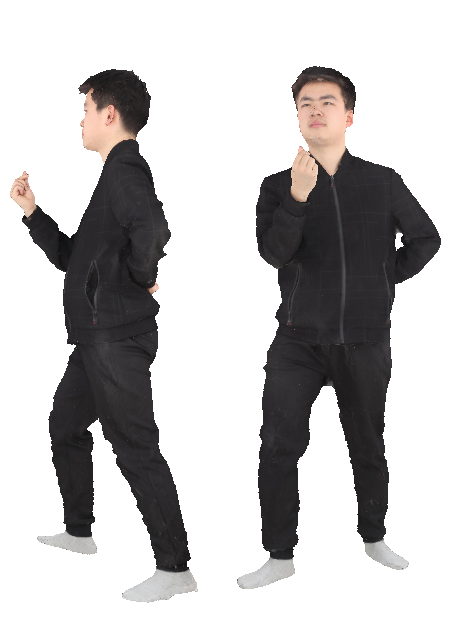}};
        \node[image node, right=1.0cm of alt2] (alt3) {\includegraphics[width=3.2cm]{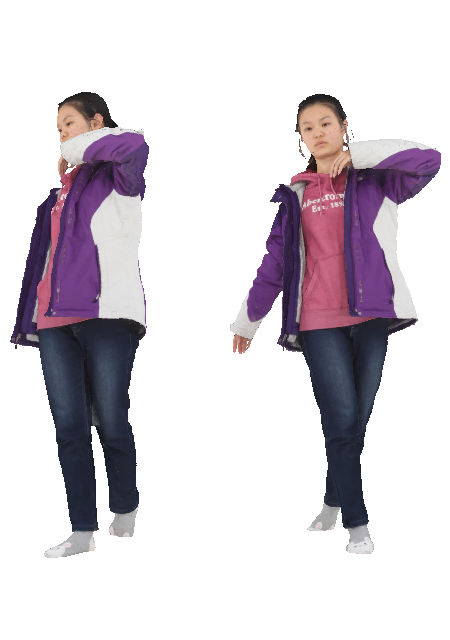}};

        \begin{scope}[on background layer]
            \fill[green!5, rounded corners=8pt] 
                ([xshift=-0.5cm, yshift=0.3cm]alt0.north west) 
                rectangle 
                ([xshift=0.5cm, yshift=-1.0cm]alt3.south east);
        \end{scope}

        \node[above=0.3cm of step2, 
              font=\footnotesize\bfseries, 
              text=blue!70!black] {(a) Iterative Orientation Adjustment Process};
        
        \path (alt0) -- (alt3) coordinate[midway] (midBottom);
        \node[above=2.25cm of midBottom, 
              font=\footnotesize\bfseries, 
              text=green!70!black] {(b) Additional Correction Examples};

        \foreach \i/\opacity in {0/0.2, 1/0.4, 2/0.6, 3/0.8, 4/1.0} {
            \fill[blue!70!black, opacity=\opacity] 
                ([yshift=-0.05cm]step\i.south) circle (1.5pt);
        }
        
        \begin{scope}[on background layer]
            \fill[blue!5, rounded corners=8pt] 
                ([xshift=-0.5cm, yshift=0.8cm] step0.north west) 
                rectangle 
                ([xshift=0.5cm, yshift=-1.2cm] step4.south east);
        \end{scope}
    \end{tikzpicture}%
    }
    \caption*{\textit{Figure 3.2: Human orientation adjustment using iterative pose estimation. \textbf{(a)} The initial model orientation is progressively corrected through iterative adjustments, where each step applies YOLOv8 pose estimation to evaluate geometric criteria and determine corrective rotations. \textbf{(b)} Additional examples showing before and after poses, demonstrating the robustness of the correction method across different initial orientations.}}
    \label{fig:orientation-adjust}
\end{figure*}
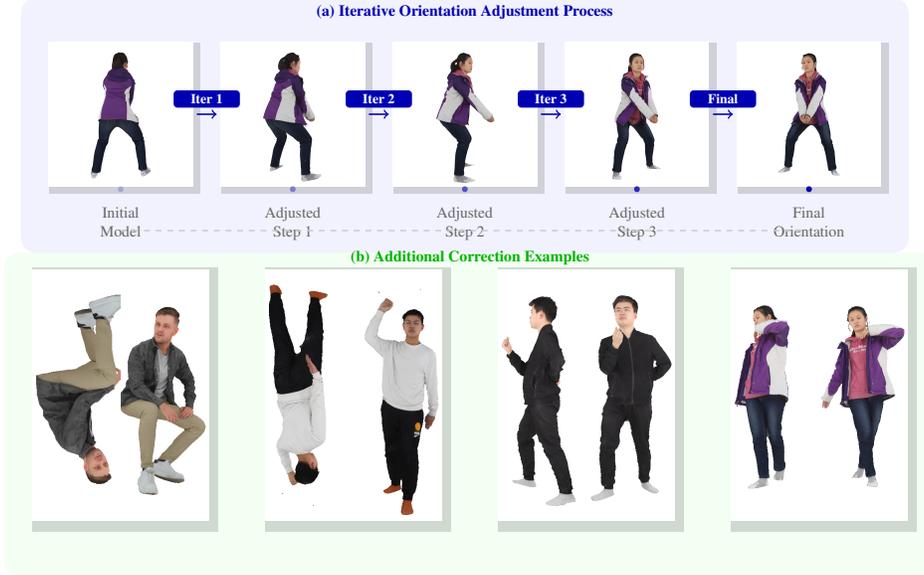

\subsection{3D Human Parsing}
A relatively new sub-field in computer vision, 3D human parsing aims to assign body part labels to point cloud or mesh vertices in synthetic data as well as real world scenes obtained with depth sensing technology. Takmaz et al. present Human3D \cite{takmaz2023human3d}, a transformer-based model designed for joint 3D human semantic segmentation, instance segmentation, and body-part segmentation in cluttered indoor scenes. Addressing the data scarcity challenge, the authors introduce a synthetic dataset generation pipeline that populates real-world ScanNet \cite{dai2017scannet} scenes with SMPL-X human meshes posed using the PLACE algorithm \cite{SMPL-X:2019}. These scenes are rendered with simulated Kinect-like noise and backprojected into point clouds with accurate ground truth labels for human instances and body parts. For real-world evaluation, the model is also trained and validated on the Behave and EgoBody datasets \cite{bhatnagar22behave, zhang2022egobody} using SMPL-X mesh fittings.
\par Human3D uses a sparse Minkowski U-Net backbone \cite{choy2019minkowski} to extract multiscale point features, and a transformer decoder that refines a set of learnable queries representing human instances and their body parts. The model applies a two-stream mask module to produce both instance and semantic segmentation masks, and uses a two-stage Hungarian matching to compute the segmentation loss. A query refinement module updates the queries based on masked cross-attention with point features and prior mask predictions. The final output is formed by merging body-part predictions into human instances using confidence thresholds and spatial masking.
\par Suzuki et al. introduce an open-vocabulary segmentation approach for 3D human models \cite{suzuki2025open}, aiming to segment point-based 3D human shapes into semantically meaningful parts based solely on text prompts. The method leverages the Segment Anything Model (SAM) \cite{kirillov2023segment} to generate multi-view, class-agnostic 2D masks from rendered images of a 3D human. These masks are back-projected back into 3D to form mask proposals. A novel HumanCLIP model is proposed to generate aligned embeddings for both visual and textual inputs, improving performance on human-centric content compared to the standard CLIP approach \cite{radford2021learning}.
\par To classify and fuse the mask proposals, a lightweight MaskFusion module is introduced, which performs cosine similarity-based matching between 3D mask embeddings and text prompt embeddings. The final segmentation is obtained by a weighted average over the 3D masks using these classification scores. This decoupling of text and mask proposals enables efficient per-prompt inference. 
\par Differing from the aforementioned approaches, this work does not use parametric mesh model body part labels, opting for the use of back-projected labels from 2D human parsing models trained on the CIHP and Humans-300 datasets. Back-projection is used to obtain ground truth labels for the ultimate aim of \textit{segmenting directly in untextured 3D space}.

\section{Method}
An overview of the proposed pipeline for obtaining pseudo ground truth labels in Figure 3.1. Mesh re-orientation is detailed in Section 3.1, follwed by a description of the back-projection to 3D from 2D human parsing approaches in Section 3.2. The downsampling stage and overall segmentation approach in 3D is given in Section 3.3.

\subsection{Mesh Orientation Alignment}
It is a common issue when using meshes and point clouds obtained from the internet that they do not follow a standard canonical orientation. A canonical orientation for a single person is defined using the following vectors:

\begin{itemize}
    \item $x_{\text{HO}} :=$ the vector from the left shoulder to the right shoulder
    \item $y_{\text{HO}} :=$ the vector from the middle of the hips to the middle of the upper chest
    \item $z_{\text{HO}} := x_{\text{HO}} \times y_{\text{HO}}$
\end{itemize}

Note that in practice vectors $x_{\text{HO}}, y_{\text{HO}}$ are not orthogonal, as required by a full world-space alignment, however one can conceptually treat the desired output as a Kabsch Rotation \cite{lawrence2019kabsch}, i.e. a world space rotation $R$ that minimizes \begin{equation} L_{align} = ||R \cdot [x, y, z]^T - [x_{\text{HO}}, y_{\text{HO}}, z_{\text{HO}}]||_{2}\end{equation} The person's eyes should also face the camera.

Let $V$ be the vertices of the mesh considered. Initially, $V$ is mean centered to $\bar{V}$, wherein the eigenvectors of the covariance matrix $Cov(\bar{V}) = \bar{V}^T \bar{V}$ are computed, sorting the eigenvectors $e_1, e_2, e_3$ by the magnitude of the eigenvalues $\lambda_1, \lambda_2, \lambda_3$. Using Rodrigues formula, the world space $y_{\text{HO}}$ is aligned with $e_1$ with the assumption that for most models, the axis of maximal variance aligns with $y_{\text{HO}}$. 
\par However, in practice this may not be the case. Consider the scenario in which a person's axis of maximum variance is more along the world's $x$ or $z$ axes, for instance if a person is lying down, then the preceding alignment may fail. 
\par To this end, a pose estimator can be used to remedy this issue. The Yolov8 \cite{yolov8} pose estimation model was used for this task, trained on the MS COCO keypoints dataset \cite{lin2014microsoft} using very heavy data augmentations for rotation in order to develop a model robust enough to predict keypoints at unusual body orientations. Using this estimator an iterative approach is developed to correct the model's orientation using keypoint locations and confidence values, computed after the initial alignment described above, as visualized in Figure 3.2.

The iterative alignment procedure operates by first rendering the mesh from a canonical viewpoint and extracting keypoint predictions using the trained pose estimator. The algorithm evaluates several geometric criteria to determine if further adjustment is necessary: (1) vertical orientation by comparing relative shoulder and hip positions, (2) depth orientation based on facial feature visibility (particularly nose confidence below $0.3$ indicating a rear-facing pose), (3) lateral lean detection through shoulder confidence asymmetry exceeding $0.15$, and (4) forward/backward lean assessment using eye confidence differentials. 
\par When these conditions indicate misalignment, targeted rotations are applied: $\pi/2$ radians about the $z$-axis for horizontal corrections, $y$-axis sign inversion for vertical flips, $z$-axis sign inversion for depth corrections, and $\pm\pi/8$ or $\pm\pi/4$ radian rotations about the $y$-axis for lean adjustments based on relative keypoint confidences. The process iterates until either the pose metrics indicate proper alignment or a maximum iteration limit is reached, with pose improvement measured primarily through ear keypoint confidence enhancement above a threshold of $0.4$, ensuring convergence toward the desired orientation where the person faces the camera in an upright stance.

\subsection{Human Parsing and Backprojection}
It is now assumed that the input mesh $M = \{V, E \}$ is properly aligned. The next step in the proposed approach is to compute human parsing segmentations for a set of views $v_i$ using pre-trained human parsing models. For each $v_i$, a parsing model is used which extracts a label image of size $I_i \in Z_{\geq 0} ^{W \times H}$, with labels from a set $L = \{l_1, l_2, ..., l_L\}$ with $l_1=0$ typically given to mean the \textit{background} class.
Views are sampled from elevations of $\{0, 30, -30\}$ and azimuth angles of $\{0, 90, 180, 270\}$, both in degrees.
\par A subtlety emerges when considering backprojection algorithms in general where projecting mesh vertices without considering their corresponding triangles leaves holes in the resulting image, which is out of distribution for human parsing models trained on standard color images. Hence it is necessary to render the mesh with standard rasterization of polygonal meshes, while determining the vertices of the relevant triangle for each non-background pixel, filtering duplicate projections by depth buffering. 
\par In order to map triangle vertices in 3D to labels in 2D, a special modification of the standard shaders of 3D rendering was used, computing a triangle ID buffer image, with shader code listed in Listings 3.1-3.2, where the \textit{evoking vertex} convention of OpenGL is used, using the last vertex of the triangle (in draw order) with the \textit{flat} keyword.  Per-point labels are then computed using a  voting procedure, using the label most voted as the final choice.

\begin{lstlisting}[numbers=none, style=glsl, caption={\textit{Listing 3.1: Vertex shader for triangle ID encoding}}, label={lst:visibility-v}]
in vec3 position;
in float vertexID; 

flat out float triangleID;

uniform mat4 model;
uniform mat4 view;
uniform mat4 projection;

void main() {
    gl_Position = projection * view * model * vec4(position, 1.0);
    triangleID = vertexID; 
}
\end{lstlisting}

\begin{lstlisting}[style=glsl, caption={\textit{Listing 3.2: Fragment shader for triangle ID encoding}}, label={lst:visibility-f}]
flat in float triangleID;   
out float fragColor;

void main() {
    fragColor = float(triangleID);
}
\end{lstlisting}

\par Following the voting procedure, the resulting vertex labels often contain noise and small disconnected components due to occlusions, rendering artifacts, and inconsistencies across views. To address this, a density-based spatial clustering approach using DBSCAN \cite{ester1996density} is applied to refine the semantic assignments. For each semantic label class $l_j \in L$, vertices assigned to that class are extracted and subjected to DBSCAN clustering with parameters $\epsilon = 0.03$ and $\text{min\_samples} = 100$. The algorithm identifies the largest cluster as the primary component for each semantic class, while smaller clusters and noise points (vertices not belonging to any dense cluster) are relabeled using k-nearest neighbor (k-NN) propagation with $k = 40$ neighbors from the remaining vertices. This denoising process effectively removes spurious label assignments and ensures spatial coherence in the final semantic segmentation, particularly important for preserving the anatomical structure of human body parts where disconnected components are typically undesirable. Further, a set of handcrafted rules are used per set of labels according to commonly observed artifacts. An example of the backprojection is shown in Figure 3.3.
\begin{figure}[t]
\centering
\begin{tikzpicture}[x=1.1cm, y=1.5cm]

% --- Row 1: Rendered Views 1–6 ---
\node[anchor=west] at (0, 0.5) {\textbf{Rendered Views}};
\foreach \i in {0,...,5} {
    \node[inner sep=0pt] at (\i, 0) {\includegraphics[width=1cm]{color_im_\i.png}};
}

% --- Row 2: Rendered Views 7–12 ---
\foreach \i [evaluate=\i as \x using \i-6] in {6,...,11} {
    \node[inner sep=0pt] at (\x, -0.8) {\includegraphics[width=1cm]{color_im_\i.png}};
}

% --- Row 3: Segmented Views 1–6 ---
\node[anchor=west] at (0, -1.6) {\textbf{Part Segmentation Views}};
\foreach \i in {0,...,5} {
    \node[inner sep=0pt] at (\i, -2.4) {\includegraphics[width=1cm]{seg_im_\i.png}};
}

% --- Row 4: Segmented Views 7–12 ---
\foreach \i [evaluate=\i as \x using \i-6] in {6,...,11} {
    \node[inner sep=0pt] at (\x, -3.2) {\includegraphics[width=1cm]{seg_im_\i.png}};
}

% --- Final Result ---
\node[anchor=west] at (0, -4.0) {\textbf{Segmented Point Cloud}};
\node[inner sep=0pt] at (2.75, -5.0) {\includegraphics[width=1.5cm, height=2.5cm]{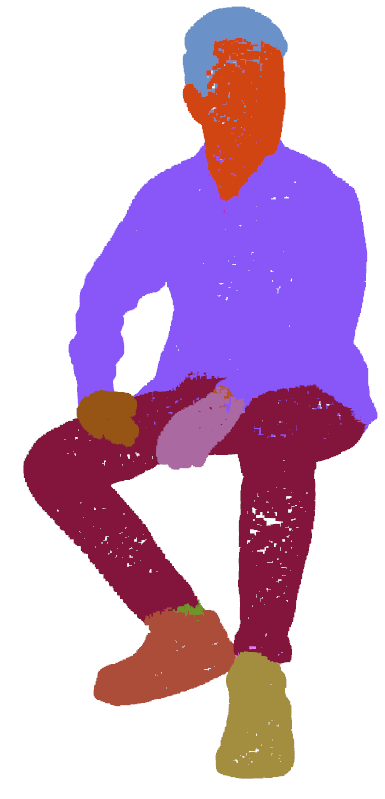}};

\end{tikzpicture}
\caption*{\textit{Figure 3.3: From rendered views and segmentation masks to the final 3D segmented point cloud. Note a subset of views is included for the purpose of clarity of presentation.}}
\end{figure}

\subsection{Subsampling and Point Cloud Parsing Segmentation}
In this section, it is assumed that there is a given dataset of point clouds with ground truth parsing labels $g_j$ for each vertices $n_j \in V$, for $N$ total points. For efficiency of preprocessing, training and inference, a novel sub-sampling approach will be employed, consisting first of Morton sorting (also referred to as serialization), followed by a \textit{windowed farthest point sampling algorithm}. For each point cloud, the values are initially normalized to $[0, 1]^3$ using coordinate-wise min-max normalization, yielding \begin{equation} n'_j \in \mathbb{R}^3 = [0.x_1 x_2 \ldots x_n, 0.y_1y_2 \ldots y_n, 0.z_1z_2 \ldots z_n]^T \end{equation} An interleaving is computed, yielding a Morton code \begin{equation} m_j = x_1y_1z_1x_2y_1z_2 \ldots x_ny_nz_n \end{equation} (removing left-trailing zeros). Points are sorted according to their Morton code and organized in $W$ windows of adjacent points, using padding of points with values at $[0, 0, 0]^T$ if necessary. Using the Morton approach, each window is locally coherent according to the properties of elementary space-filling curves, and by reducing the iterative farthest point sampling operation \cite{qi2017pointnetplusplus}, for $k$ samples to $k/W$ samples for each of $k$ windows, run time is reduced from $O(kN) $ to $O(k N/W)$. Window size is chosen using a maximum number of points per window of $5000$, where the total number of windows is \begin{equation} (N + 5000 - 1) \pmod {5000} \end{equation} Further, part-specific oversampling is used for body parts with less samples as an attempt to ensure high mIoU, such as the hands, face, arms and hair, depending on the parsing labels which may have less point counts. An example sub-sampling is shown in Figure 3.4.
\begin{figure}[t]
  \centering

  \begin{subfigure}[b]{0.3\textwidth}
    \includegraphics[width=\linewidth]{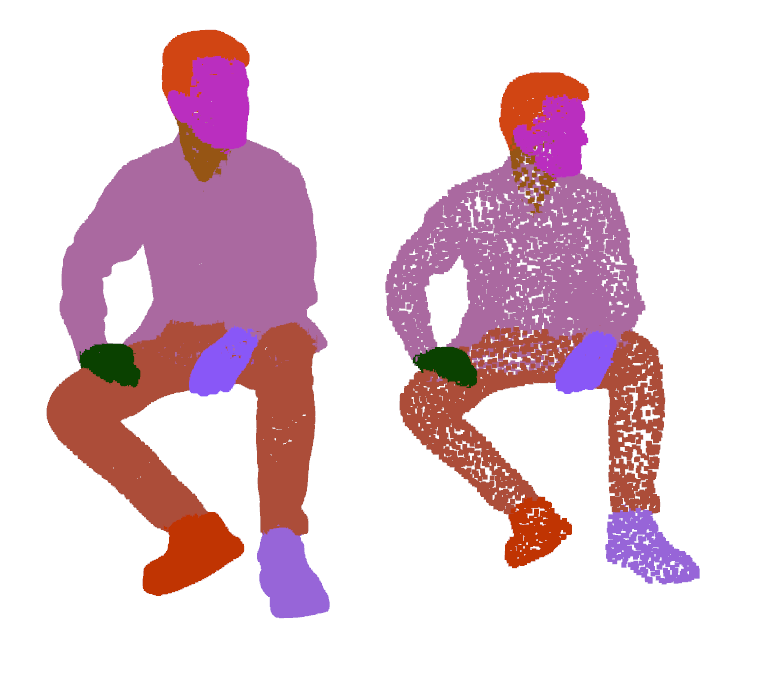}
  \end{subfigure}

  \vspace{0.5em}

  \begin{subfigure}[b]{0.3\textwidth}
    \includegraphics[width=\linewidth]{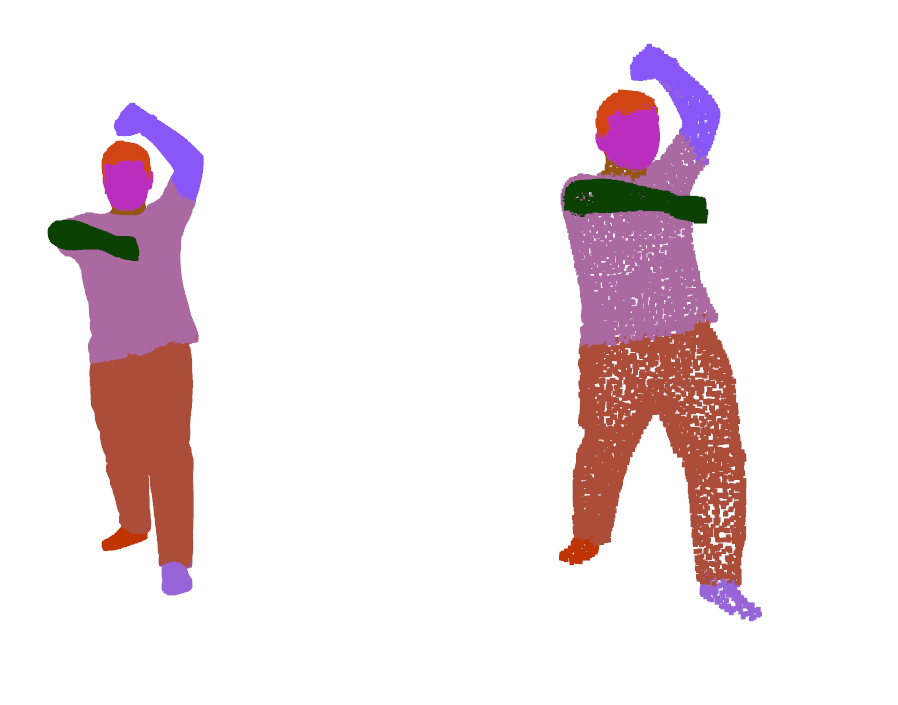}
  \end{subfigure}

  \caption*{\textit{Figure 3.4: Visualizing the subsampling of human point clouds. 
  Arm, face, hair, and feet points are sampled more frequently. 
  For each row, on the left the original model is shown, and on the right 10,000 sampled points are displayed.}}
  \label{fig:stacked_images}
\end{figure}

For point cloud segmentation, the PointTransformer model \cite{zhao2021point}, a relatively simple yet high-performing approach by modern standards was selected as a reasonable baseline, though the specific choice of model remains agnostic, where for full detail the reader is encouraged to read the original publication. 
\par The PointTransformer semantic segmentation model is organized into $11$ blocks of layers. Using K-NN as a feature aggregator, where for a point $x_i$, its features are computed as \begin{equation}
\sum_{x_j \in N(x_j)} \rho(\gamma(\phi(x_i) - \psi(x_j) + \delta)) \odot (\alpha(x_j) + \delta) 
\end{equation}
where $\gamma, \phi, \psi,  \alpha$ are point-wise features computed by an MLP (Multi-Layer Perceptron), $\rho$ is the softmax function, and $\delta$ are positional encodings given by MLP features on the difference in coordinates of neighbors to the central point, where point-wise neighbors are denoted as $N(x_j)$. 
\par At the beginning of blocks 2-6, in the initial layer per block, iterative farthest point sampling yields a subset $P2 \subset P1$, neighbors from $P1$ are used for feature aggregation, referred to as the \textit{transition down} operation. After the backbone layers have computed features and downsampled the input point to a set of $N/S$ for stride parameter $S$, point-features are gradually upsampled with a \textit{transition up} operation for blocks 7-10 using trilinear interpolation and point-wise MLP learning for points not sampled, with the final layer mapping each point to a class label.

\section{Experiments}
\subsection{Dataset}
Experiments were conducted on the THuman2.1 dataset \cite{tao2021function4d}, a comprehensive collection of high-quality 3D human scans, where THuman2.1 is an extension of the THuman2.0 release. Each model is captured by a calibrated 128-camera DSLR multi-view rig, yielding a very dense 3D mesh (on the order of hundreds of thousands to millions of vertices) along with aligned texture maps, resulting in 2,445 scans with full texture, providing a balance of properties essential for the proposed two-stage approach. The detailed geometry captured through dense multi-view reconstruction preserves fine-scale surface features including clothing wrinkles, fabric textures, and anatomical details necessary for geometry-based segmentation learning. Moreover, the high-quality texture maps enable reliable pseudo-ground truth generation through 2D human parsing algorithms.

\subsection{Implementation Details}
For human parsing, two models with corresponding label sets were considered, namely Mask2FormingParsing (M2FP) \cite{yang2023humanparsing}, using the Crowd Instance Human Parsing (CIHP) label set, and the Sapiens \cite{khirodkar2024sapiens} model with a newer set of labels, which we restrict to subsets of label spaces shown in Table 4.1. 

\begin{table}[h!]
  \centering
  \label{tab:example}
  \begin{tabular}{l p{5cm}}  % left column normal, right column wraps at 10cm
    \toprule
    Model  & Labels \\
    \midrule
    M2FP (CIHP) & background, hat, hair, gloves, sunglasses, upper clothes, dress, coat, socks, pants, torso-skin, scarf, skirt, face, left arm, right arm, left leg, right leg, left shoe, right shoe \\
    Sapiens v1 & background, apparel, face and neck, hair, left foot, left hand, left arm, left leg, lower clothing, right foot, right hand, right arm, right leg, torse, upper clothing \\
    Sapiens v2 & background, apparel, face and neck, hair, left foot, left hand, left arm, left leg, lower clothing, right foot, right hand, right arm, right leg, torse, upper clothing, lip, teeth, tongue \\
    \bottomrule
  \end{tabular}
  \caption*{\textit{Table 4.1: The proposed label spaces, note that not all instance of each label appear in the ground truth. Sapiens v2 is an extension of the first label space where lip, teeth, and tongue labels were included.}}
\end{table}
The dataset was randomly shuffled and manually split into training, validation, and testing sets, following a 75/10/15 ratio (train, val, test). All experiments were conducted in PyTorch using two Nvidia 4090 GPUs. Two different sampling sizes were used in the proposed windowed iterative farthest point sampling, 10,000 and 100,000 points, using batch sizes of 32 and 6 respectively. Experiments with oversampling certain parsing labels were performed to see if performance can be enhanced, namely the arms face and hair for the CIHP labels, and the hands, arms, lip, teeth and tongue for the Sapiens label set.
\par The SGD (Stochastic Gradient Descent) optimizer was used with a base learning rate of 0.01 using a cosine annealing schedule, with Nesterov momentum and a weight decay of $10e^{-5}$ for 100 epochs. The loss function was an equally weighted sum of cross-entropy and the Dice loss, as is now common in semantic segmentation models \cite{cheng2021maskformer, cheng2021mask2former}. Data augmentation consisted of point jitter and random rotations about the world-space y-axis. 
\subsection{Results}
The segmentation results for the proposed experiments is shown in Table 4.2, where three different metrics to evaluate model performance are considered, namely
let:
\begin{itemize}
  \item \( n_{ij} \): number of points of class \( i \) predicted as class \( j \)
  \item \( n_{ii} \): number of true positives for class \( i \)
  \item \( t_i = \sum_j n_{ij} \): total number of points of class \( i \)
  \item \( n_{\text{classes}} \): total number of classes
  \item \( N = \sum_i t_i \): total number of points
\end{itemize}

\begin{equation}
\text{\textit{mIou}}= \frac{1}{n_{\text{classes}}} \sum_{i=1}^{n_{\text{classes}}} \frac{n_{ii}}{t_i + \sum_{j} n_{ji} - n_{ii}}
\end{equation}

\begin{equation}
\text{\textit{fw mIou}} = \frac{1}{N} \sum_{i=1}^{n_{\text{classes}}} t_i \cdot \frac{n_{ii}}{t_i + \sum_{j} n_{ji} - n_{ii}}
\end{equation}

\begin{equation}
\text{\textit{Acc}} = \frac{\sum_{i} n_{ii}}{N}
\end{equation}

\noindent where \textit{mIoU} denotes mean intersection over union, \textit{Acc} denotes accuracy, and \textit{fw} denotes frequency weighted.  Qualitative results for parsing are shown in Figure 4.1.

\begin{figure}[h!]
    \centering

    \includegraphics[width=0.8\linewidth]{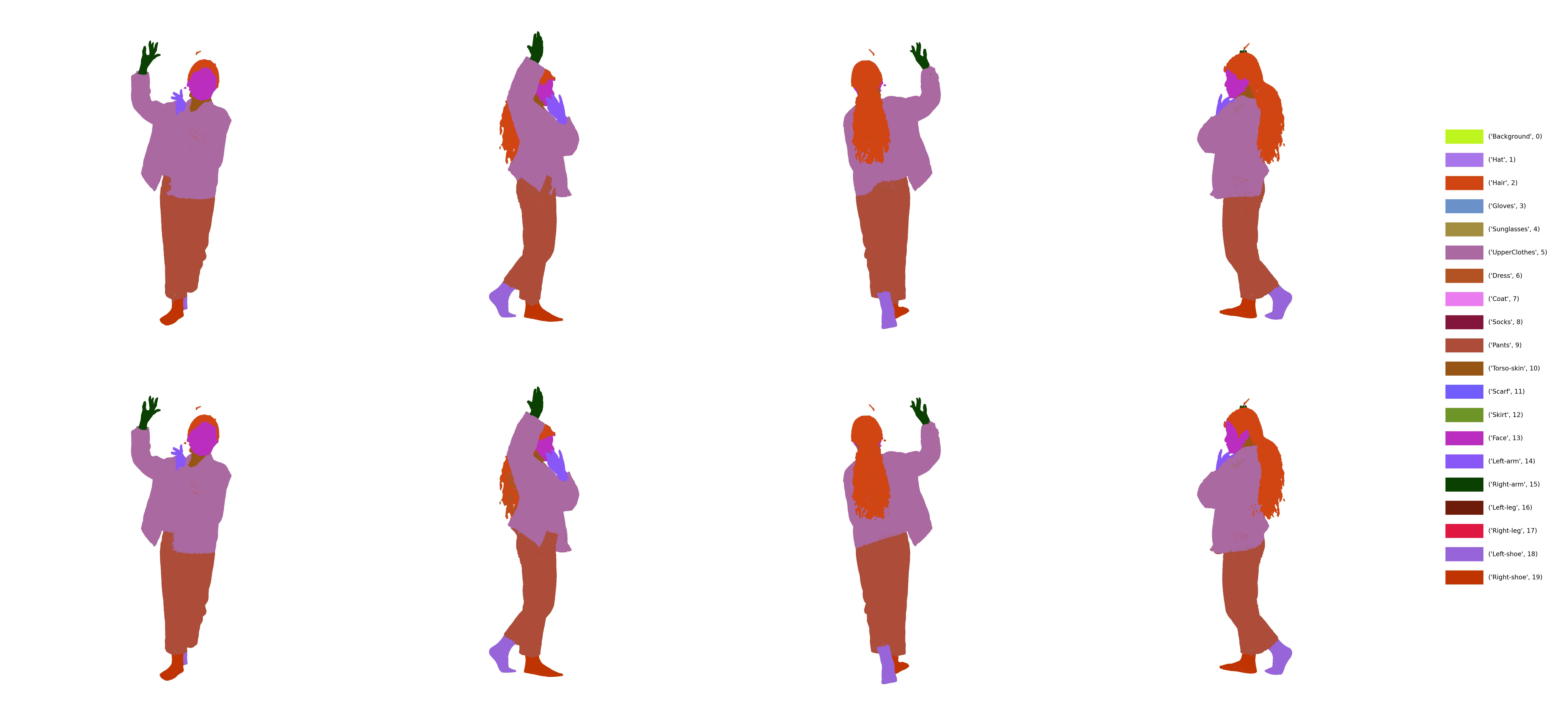}

    \includegraphics[width=0.8\linewidth]{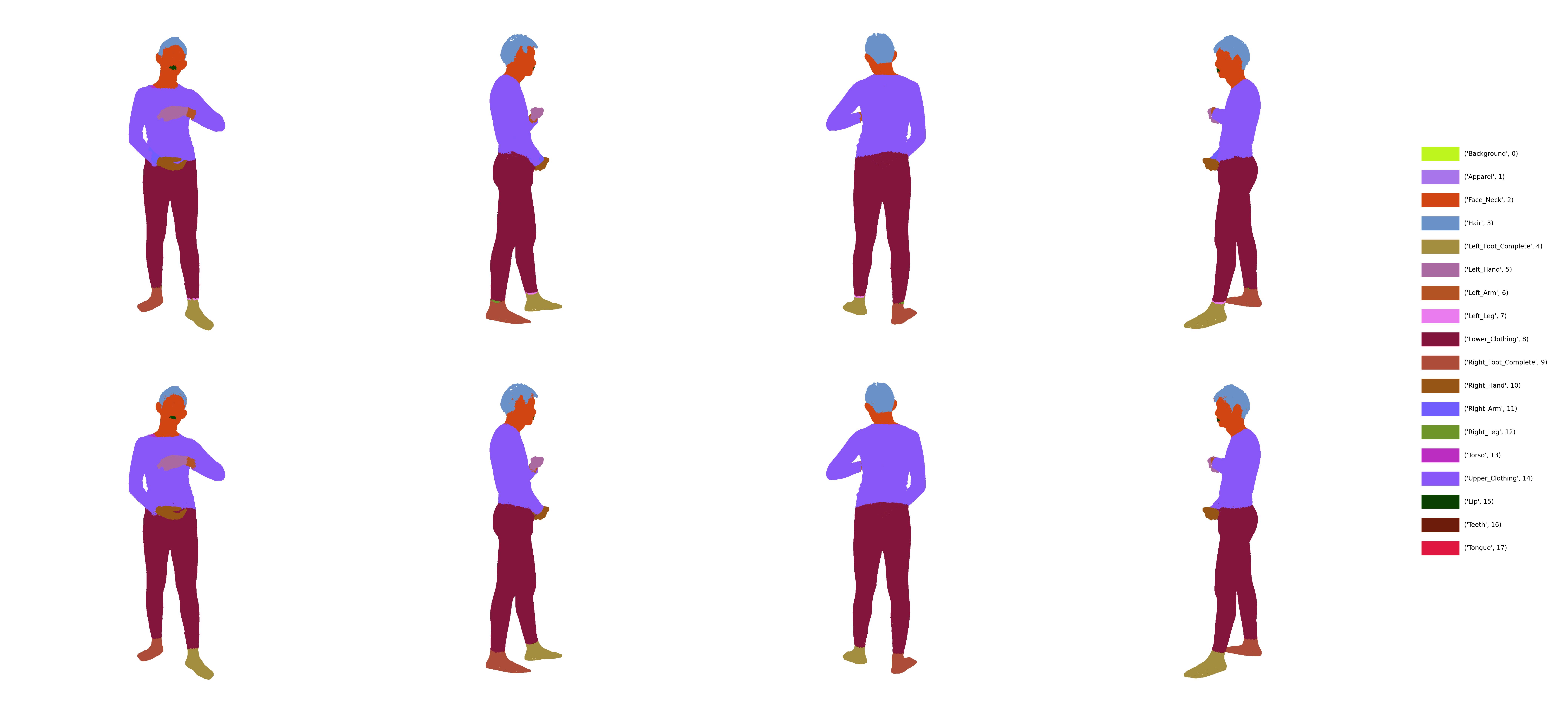}

    \label{fig:stacked_images}
    \caption*{\textit{Figure 4.1: Qualitative examples of segmentation results with the proposed approach, viewed with 4 viewing angles with the ground truth displayed above, and the predictions displayed below. Top, an example using CIHP lables, with ground truth obtained from the M2FP model. Bottom, an example using Sapiens v2 labels obtained with the Sapiens model.}}
\end{figure}

\begin{table}[h!]
  \centering
  \label{tab:example}
  \begin{tabular}{lccc}
    \toprule
    Labels and \# of Points & \textit{mIoU} & \textit{fw mIoU} & \textit{Acc}\\
    \midrule
    % Add rows like:
    CIHP, 10K points & 54.8 & 87.2 & 92.9 \\
    CIHP-O, 10K points & 56.1 & 87.8 & 93.3 \\
    CIHP, 100K points & 60.7 & 93.3 & 96.4 \\
    CIHP-O, 100K points & 58.7 & 93.3 & 96.4 \\
    Sapiensv1, 10K points & 58.8 & 87.3 & 93.0 \\
    Sapiensv1-O, 10K points & 59.5 & 87.6 & 93.1\\
    Sapiensv1, 100K points & 74.4 & 93.9 & 96.8 \\
    Sapiensv1-O, 100K points & 67.9 & 91.1 & 95.1 \\
    Sapiensv2, 10K points & 52.5 & 88.7 & 93.8\\
    Sapiensv2-O, 10K points & 47.1 & 84.1 & 90.6\\
    Sapiensv2, 100K points & 65.3  & 94.0  & 96.8 \\
    Sapiensv2-O, 100K points & 59.9  & 91.8 & 95.5 \\
    \bottomrule
  \end{tabular}
    \caption*{\textit{Table 4.2: Semantic segmentation results of the PointTransformer model on the CIHP and Sapiens labeled data, using 10,000 and 100,000 points, where O denotes oversampled part labels, see Section 4.2. Results are reported after the final nearest neighbor upsampling using 3 neighbors. All values are in percentages.}}
\end{table}
\section{Discussion}
In general, the results shown in Section 4.3 demonstrate the effectiveness of the proposed approach, with some caveats. In particular, with respect to early experiments, the mIoU metric of model predictions in the original number of sampled points (10,000, and 100, 000) was much higher before upsampling with nearest neighbors to the full original vertex count of meshes considered, which is a common issue noted in semantic segmentation as noted by Long et al. \cite{long2015fully}. This suggests either the development of a model capable of predicting even larger numbers of point labels, or a more sophisticated upsampling approach, which is left to future research, although it is noted that the point sampling values 10,000 and 100,000 are an order of magnitude larger than the typical 2048 points used in point cloud part segmentation literature \cite{chang2015shapenet}. 
On the whole models trained with 100,000 points outperformed those with 10,000 points. Further in the 10,000 point regime, oversampling parts resulted in a slight boost in all metrics, but did not in the 100,000 point regime.

\par In analyzing the weakest performing classes, the label sets used reveals areas for future improvement. In particular:
\begin{itemize}
\item For the \textbf{CIHP} labeled meshes, the weakest performing classes were the legs, torso skin, and the dress class, partly owing to the ambiguity of labeling clothing.
\item For the \textbf{Sapiens v1} labeled meshes, the weakest performing classes were the torso, apparel, and legs and arms classes.
\item For the \textbf{Sapienvs v2} labeled meshes, the same issues as v1 occurred, except for the lip class being also an issue, explained by its low point count in the training data.
\end{itemize}

In general, challenging areas tend to involve parts with fewer points in the training data, as well as transitional regions (e.g., between legs and shoes). The former can be attributed to limited label coverage in datasets like CIHP and Sapiens, which offer a restricted set of clothing labels that struggle to capture the wide variability of real-world attire. The latter is often due to labeling ambiguities at region boundaries, making precise segmentation difficult.
\section{Conclusion}
In conclusion, in this work, a method to obtain ground truth per-point labels from human meshes was developed, leveraging techniques from human parsing as well as point cloud research within a deep learning framework. The resulting model can input pure geometry, i.e. a point cloud without texture or surface normal information, and predict human parsing labels related to clothing and body parts. 
\par Future research into this topic should largely focus on two main constituents. Firstly, the label space used for a general parsing approach should be as wide as possible, incorporating many clothing types and accessories that are not featured in the CIHP and Sapiens labels. This lack of label diversity can cause out of distribution issues in the closed-set model setting. Secondly, while the proposed approach can operate at inference time with under a second speed for an NVidia 4090 gpu ($ \approx 0.73$ seconds), this may still be too slow for certain applications. Morevoer, the number of views used as well as the use of the cpu-bound DBSCAN algorithm in the process of obtaining pseudo ground truth labels is slow and difficult to parallelize, hence more efficient methods should be pursued to speed up the overall pipeline. 
\clearpage
\bibliographystyle{ieee}
\bibliography{refs}

\end{document}